\documentclass{article}

\usepackage[preprint]{neurips_2022}

\setcitestyle{authoryear, open={(},close={)}}

\usepackage[utf8]{inputenc} 
\usepackage[T1]{fontenc}    
\usepackage{hyperref}       
\usepackage{url}            
\usepackage{booktabs}       
\usepackage{amsfonts}       
\usepackage{nicefrac}       
\usepackage{microtype}      
\usepackage{xcolor}         
\usepackage{multirow}
\usepackage[pdftex]{graphicx}
\usepackage{wrapfig}
\usepackage{amsmath}
\usepackage{diagbox}

\title{Advanced Feature Learning on Point Clouds using Multi-resolution Features and Learnable Pooling}

\author{%
 Kevin Tirta Wijaya$^1$\thanks{co-first authors} \quad\quad  Dong-Hee Paek$^{2*}$ \quad\quad Seung-Hyun Kong$^2$\thanks{corresponding author} \\
  $^1$Robotics Program \quad $^2$CCS Graduate School of Mobility \\
  KAIST\\
  \texttt{\{kevin.tirta, donghee.paek, skong\}@kaist.ac.kr} \\
}

\begin{document}

\maketitle

\begin{abstract}
Existing point cloud feature learning networks often incorporate sequences of sampling, neighborhood grouping, neighborhood-wise feature learning, and feature aggregation to learn high-semantic point features that represent the global context of a point cloud.
Unfortunately, such a process may result in a substantial loss of granular information due to the sampling operation.
Moreover, the widely-used max-pooling feature aggregation may exacerbate the loss since it completely neglects information from non-maximum point features.
Due to the compounded loss of information concerning granularity and non-maximum point features, the resulting high-semantic point features from existing networks could be insufficient to represent the local context of a point cloud, which in turn may hinder the network in distinguishing fine shapes.
To cope with this problem, we propose a novel point cloud feature learning network, PointStack, using multi-resolution feature learning and learnable pooling (LP).
The multi-resolution feature learning is realized by aggregating point features of various resolutions in the multiple layers, so that the final point features contain both high-semantic and high-resolution information.
On the other hand, the LP is used as a generalized pooling function that calculates the weighted sum of multi-resolution point features through the attention mechanism with learnable queries, in order to extract all possible information from all available point features.
Consequently, PointStack is capable of extracting high-semantic point features with minimal loss of information concerning granularity and non-maximum point features.
Therefore, the final aggregated point features can effectively represent both global and local contexts of a point cloud.
In addition, both the global structure and the local shape details of a point cloud can be well comprehended by the network head, which enables PointStack to advance the state-of-the-art of feature learning on point clouds.
Specifically, PointStack outperforms various existing feature learning networks for shape classification and part segmentation on the ScanObjectNN and ShapeNetPart datasets.
The codes are available at https://github.com/kaist-avelab/PointStack.

\end{abstract}

\section{Introduction}

Point cloud has been one of the most popular representations of 3D objects in recent years \citep{qi2017pointnet_pointnet, yang20203dssd_pointbased1, yu2021pointr_pointbased2}.
The capability of point cloud data in representing highly-complex 3D objects with low memory requirements enables real-time 3D vision applications for resource-limited agents.
This is contrary to the voxel-based representations \citep{graham20183d_voxelbased1,yan2018second_voxelbased2}, where the memory requirement is cubically proportional to the spatial resolution.
In addition, point cloud is the native data representation of most 3D sensors, thus performing 3D vision directly on point clouds minimizes the pre-processing complexity.
The advantages indicate that point cloud can be the prime data representation for fast and accurate neural networks for 3D vision.

Unfortunately, there are several challenges in applying the matured 2D deep learning-based feature learning techniques to the point cloud, for example, the irregular and unordered nature of the point cloud.
These issues are addressed in the pioneering works of \citet{qi2017pointnet_pointnet} and \citet{qi2017pointnet++_pointnet++}, which present the multilayer perceptron-based (MLP-based) PointNet and PointNet++, respectively.
In the PointNet++ framework, a sequence of keypoints sampling, neighborhood grouping, neighborhood-wise feature learning, and feature aggregation is repeated several times to produce high-semantic point features.
The relatively simple framework of PointNet++ is widely-used in literature.
For example, PointMLP \citep{ma2022rethinking_pointmlp} enhances the framework by incorporating residual connections and construct a 40-layer MLP-based network that achieves state-of-the-art classification performance on several datasets.

Despite the promising results, the final high-semantic point features of the PointNet++ framework lose granular information due to the repeated keypoints sampling, where each of the surviving point features at the deeper layer of the network represents a larger spatial volume in the point cloud.
Moreover, the max-pooling function that is used for feature aggregation may exacerbate the loss since it completely neglects information from non-maximum point features.
Such a compounded loss of information concerning granularity and non-maximum point features could substantially harm the capability of the point features in delivering local context information such as the detailed shapes of objects in point clouds.

Considering the problem of losing granular and non-maximum point features information, we present two hypotheses. 
(1) It is advantageous for the task-specific head to have access to point features from all levels of resolutions.
This enables the network to extract high semantic information while preserving the granularity to a certain degree.
(2) A generalized pooling function that combines information from all point features could improve the representation capacity of the aggregated point features since the loss of information from non-maximum point features is minimized.

Based on the hypotheses, we propose a novel MLP-based network for feature learning on point clouds, PointStack, with multi-resolution feature learning and learnable pooling (LP).
PointStack collects point features from various resolutions that are already available in the multiple layers of the PointNet++.
The collected multi-resolution point features are then aggregated and fed to the task-specific head.
Therefore, the task-specific head has access to both high-semantic and high-resolution point features.
Moreover, PointStack utilizes the LP that is based on the multi-head attention (MHA) mechanism \citep{vaswani2017attention_transformer} with learnable queries for both single-resolution and multi-resolution feature aggregation.
The LP is a permutation invariant and generalized pooling function that does not neglect information from non-maximum point features, but calculates the weighted sum of the multi-resolution point features according to their attention scores.
Consequently, PointStack is capable of producing high-semantic point features with minimal loss of information concerning granularity and non-maximum point features, such that both global and local contexts of a point cloud can be effectively represented.
As a result, the network head can well comprehend the global structure and distinguish fine shapes of a point cloud, enabling PointStack to advance the state-of-the-art of feature learning from point clouds.

Specifically, we observe that PointStack exhibits superior performance compared to various existing feature learning networks on two popular tasks: shape classification that requires global context understanding, and part segmentation that requires both global and local context understanding.
On the shape classification task with ScanObjectNN dataset, PointStack outperforms existing feature learning networks by 1.5\% and 1.9\% in terms of the overall and class mean accuracy, respectively.
On the part segmentation task with ShapeNetPart dataset, PointStack outperforms existing feature learning networks by 0.4\% in terms of the instance mean intersection over union metric.
The two results demonstrate the superiority of the proposed PointStack, not only for tasks that require global context, but also for tasks that require local context.

In a summary, our contributions are as follows,
\begin{itemize}
    \item{In the proposed PointStack, we employ a multi-resolution feature learning framework for point clouds. Leveraging point features of multiple resolutions provides both high-semantic and high-resolution point features to the task-specific heads. Therefore, the task-specific heads can obtain high-semantic information without substantially losing granularity.
    }
    
    \item {We propose a permutation invariant learnable pooling (LP) for point clouds as an advancement to the widely-used max pooling.
    LP is a generalized pooling compared to the max pooling, since it combines information from multi-resolution point features through the multi-head attention mechanism, as opposed to only preserving the highest-valued features.
    }
    \item{We demonstrate that PointStack outperforms various existing feature learning networks for point clouds on two popular tasks that includes shape classification on the ScanObjectNN dataset and part segmentation on the ShapeNetPart dataset.}
\end{itemize}

The remaining of this paper is organized as follows.
Section 2 discusses existing works that are related to feature learning.
Section 3 describes the proposed PointStack in detail.
Section 4 shows the experimental results with extensive discussions.
Section 5 concludes this work with a summary.

\section{Related Work}
\textbf{Feature Learning on Point Clouds.}
Most of the modern feature learning neural networks for point cloud data originate from the pioneering work PointNet by \citet{qi2017pointnet_pointnet}.
In PointNet, a sequence of point-wise multilayer perceptron (MLP) blocks is applied to the raw point cloud to produce high-dimensional point features.
The point features are then aggregated through the max pooling operation, which results in a fixed-length global feature vector.
PointNet++ \citep{qi2017pointnet++_pointnet++} refines the PointNet by considering local structures of the point via sampling, grouping, and local group feature aggregation.
First, a collection of keypoints is obtained through farthest-point sampling.
Then, neighboring points around each keypoint are grouped, and a PointNet operation is applied to each group, resulting in a neighborhood-wise global feature vector for each keypoint.

Since then, numerous research has been conducted to enable learning the fine-grained local geometrical features of the point clouds.
For example, \citet{wang2019dynamic_dgcnn} propose a method to learn the relationships between the points with graph-based EdgeConv.
\citet{wu2019pointconv_pointconv} introduce a convolution-based network that learns the appropriate convolution kernel via MLP networks and kernel density estimation.
\citet{hamdi2021mvtn_mvtn} present a multi-view approach, where the network regresses the optimal view-point of the objects for 3D recognition.
Recently, \citet{ma2022rethinking_pointmlp} introduce PointMLP, a relatively deep MLP-based network for point cloud.
The network is based on the original PointNet++ with additional residual connections and geometric affine modules.
Owing to the residual connections, PointMLP manages to comprise deep layers, where the best-performing variant is composed of 40 layers.

\textbf{Deep Learning with Multi-resolution Features.}
Multi-resolution features have been extensively explored in the image-based computer vision.
Various traditional image processing techniques, such as the ones introduced by \citet{dalal2005histograms_hog} and \citet{lowe2004distinctive_sift}, utilize a feature pyramid that leverages features of various resolutions (scales) from multiple layers for the downstream task prediction.
The feature pyramid framework is still widely used in the deep-learning, especially after the introduction of the Feature Pyramid Network (FPN) by \citet{lin2017feature_fpn}.
In FPN, feature maps of multiple resolutions are downsampled or upsampled to match the output feature map size, and concatenated together, resulting in an output feature map with both high-resolution and high-semantic information.
In the point cloud domain, \citet{hui2021pyramid_pyramidpointtransformer} propose a transformer-based feature extractor that learns multi-scale feature maps for large-scale place recognition.

\section{PointStack: Multi-resolution Feature Learning with Learnable Pooling}
In this section, we first introduce an overview of the proposed multi-resolution feature learning implemented on a deep MLP-based network, PointStack.
Following the overview, we introduce the learnable pooling that is permutation invariant.

\subsection{Multi-resolution Feature Learning}

\begin{figure}[t]
    \centering
    \includegraphics[width=0.95\textwidth]{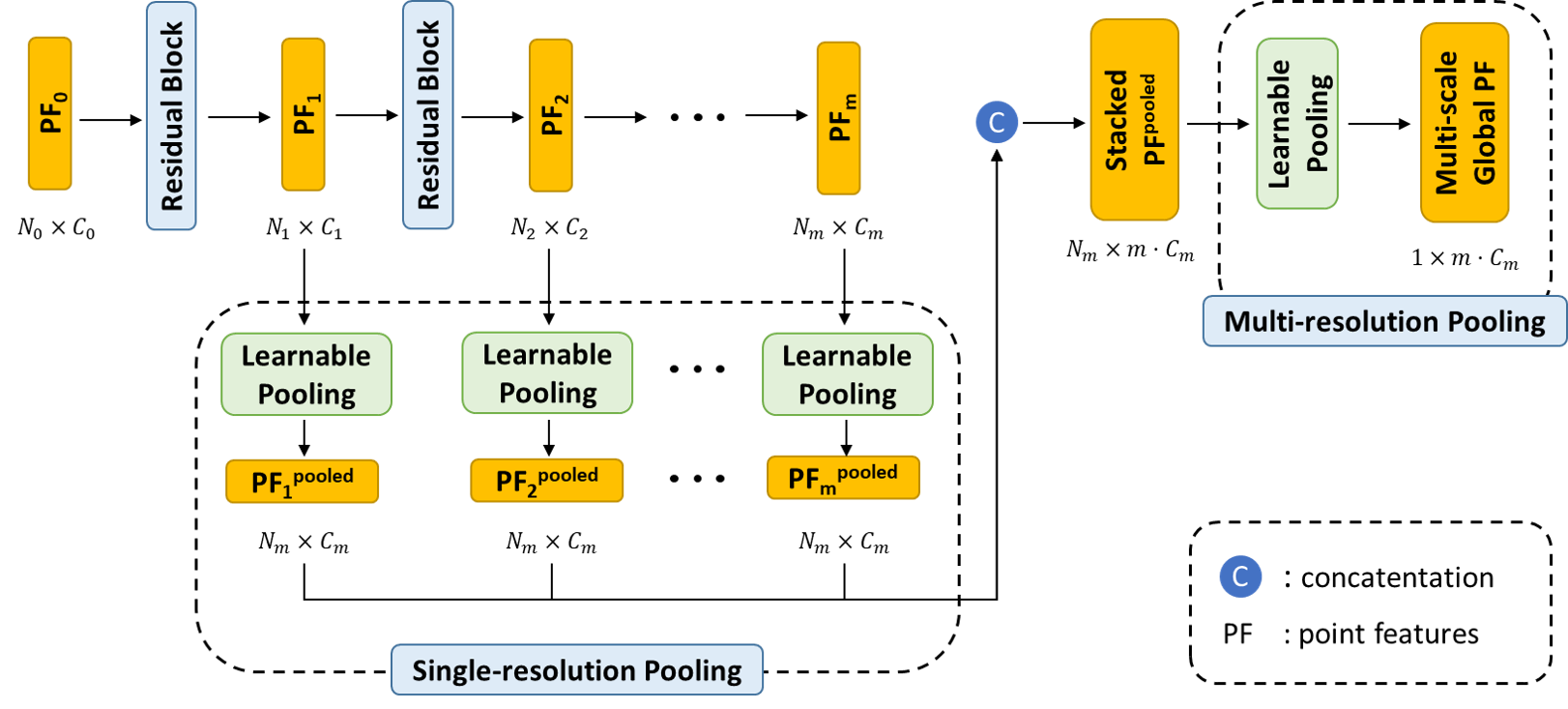}
    \caption{Feature learning backbone of PointStack. The residual block (one stage of PointMLP, \citet{ma2022rethinking_pointmlp}) learns the underlying representation of the input point features and outputs point features of a reduced length. For \textit{m} repeated residual blocks, the output point features of each block are pooled by the learnable pooling (LP) and concatenated to form stacked point feature. The final LP is then applied to obtain the multi-resolution features, which can be used by the network heads.}
    \label{fig:pointstack}
\end{figure}

The concept of multi-resolution feature learning is widely used for various downstream tasks in the computer vision \citep{lin2017feature_fpn, ghiasi2019fpn_nasfpn, kirillov2019panoptic_panopticfpn}.
The principle approach is to construct a feature pyramid from semantic features from all levels of resolutions.
As a result, the feature pyramid has both high-semantic and high-resolution information that is often needed to recognize objects of various scales.

In the 3D point cloud domain, the potential benefits of utilizing multi-resolution features arise from the fact that 3D shapes are significantly more complex compared to the 2D images.
Important textures or curves of the 3D shapes may be only observable at the highest level of granularity.
As constructing high-semantic features comes at a cost of losing granularity in the existing approaches, the finer details of the 3D shapes may be obscured.
Therefore, multi-resolution point features can be a solution for both collecting sufficient semantic information and preserving granularity to a certain degree.

Unlike PyramNet \citep{zhiheng2019pyramnet} that uses multiple convolutions with different kernel sizes on the same point features to create a multi-scale feature map, we opt to leverage multiple point features from \textit{m} different resolutions that are already available in the existing MLP-based networks, as shown in Figure \ref{fig:pointstack}.
PointStack first learns the underlying representations of the points with the \textit{m} repeated residual blocks, where the output of each block has lower resolution but higher semantic information compared to the corresponding input.
We use the residual blocks instead of the transformer blocks as in \citet{hui2021pyramid_pyramidpointtransformer}, since residual blocks are more efficient in terms of the memory requirements.
This is because the self-attention mechanism in each of the transformer blocks has $\mathcal{O}(n^2)$ memory complexity with respect to the input size \textit{n}.

After learning the appropriate representations, PointStack performs single-resolution pooling on each output point features, as shown within the bottom-left dotted box in Figure \ref{fig:pointstack}.
That is, PointStack pools from each output point features ($\textbf{PF}_i$ of $N_i$ feature vectors) at the \textit{i}-th layer to produce $\textbf{PF}_i^{pooled}$ of a fixed length $N_m$, where $\textbf{PF}^{pooled}$ contains important features on the specific level of resolution.

Following the single-resolution pooling, PointStack concatenates all $\textbf{PF}_i^{pooled}$ to form and process the stacked pooled point features, stacked-$\textbf{PF}^{pooled}$, through the multi-resolution pooling (top-right dotted box in Figure \ref{fig:pointstack}), to produce a global feature vector.
Since the global feature vector is obtained from the features of \textit{m} resolutions, it contains information from both high-semantic and high-resolution features.
Therefore, the task-specific heads have access to high-semantic information with minimal loss of granularity.

Note that the multi-resolution feature learning framework can be realized without fixing the length of the output features of the single-resolution pooling.
However, we find empirically that the fixed-length single-resolution pooling substantially improves the classification performance.
Such a phenomenon may originate from the fact that point features of \textit{m} different resolutions have different numbers of entries.
That is, the highest-resolution point feature, $\textbf{PF}_1$, has significantly more feature vectors compared to the lowest-resolution point feature, $\textbf{PF}_m$.
The disparity between the number of feature vectors may adversely affect the multi-resolution LP.
Therefore, we incorporate the single-resolution pooling process to produce the same number of feature vectors from \textit{m} resolution levels.
This explanation is supported by the experimental results shown with the ablation study in Section 4.

\subsection{Learnable Pooling}
\begin{wrapfigure}{r}{0.5\textwidth}
    \centering
    \includegraphics[width=0.5\textwidth]{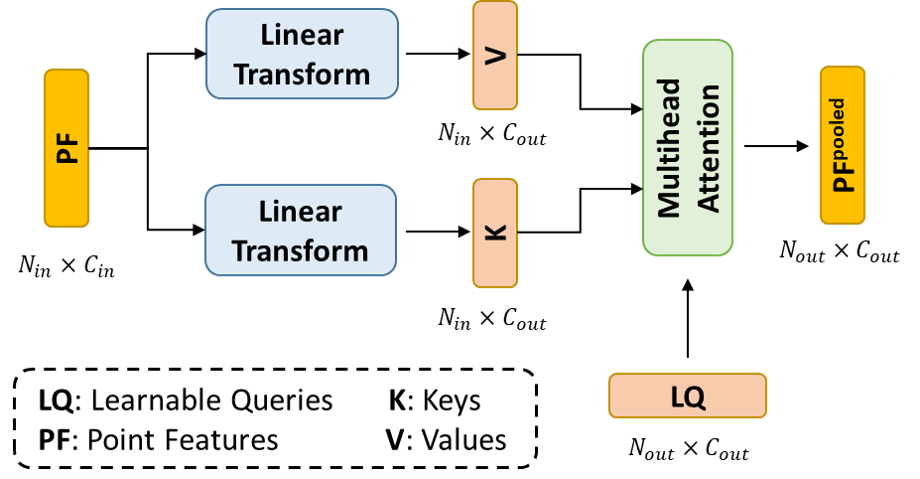}
    \caption{The structure of the learnable pooling (LP) module. Given an input of point features, LP transforms the features such that the channel size of the features match the channel size of the learnable queries (LQ). A multi-head attention (MHA) mechanism is then used to produce the fixed-length pooled point features. For the MHA, we set the input point features as the source of the \textit{keys} and \textit{values}, and LQ as the \textit{queries}.}
    \label{fig:learnable_pooling}
\end{wrapfigure}
Recent works for feature learning on point clouds often utilize pooling functions.
The pooling function is an important trick to produce fixed-length global features from input points of an arbitrary size.
Since a 3D shape can be represented by the same set of points of different order, the pooling function should be permutation invariant.
A natural choice for such requirements is the max pooling function.
Unfortunately, the max pooling function only preserves the highest-valued point features and completely neglects non-maximum point features, which results in a substantial amount of information loss.

To prevent this problem, we propose a generalized pooling function, learnable pooling (LP), that aggregates by calculating the weighted sum of all point features according to the correlation between the point features and learnable parameters.
Since the LP does not neglect information from non-maximum point features, it can be used for aggregating both single-resolution and multi-resolution point features without loss of information.

Structurally, LP utilizes the multi-head attention (MHA) \citep{vaswani2017attention_transformer} that can be seen as an information retrieval process, where a set of \textit{queries} is used to retrieve information from the \textit{values} based on the correlation between the \textit{queries} and the \textit{keys}.
We set both \textit{keys} and \textit{values} to originate from the same point feature tensor, while the \textit{queries} are learnable parameters.
In this setting, we can consider the network to learn the appropriate \textit{queries} so that the retrieved point features (\textit{values}) are highly relevant to the learning objectives.
As the \textit{queries} are directly supervised by the learning objectives, and the \textit{values} are obtained through the weighted sum of all point features, the proposed LP is capable of producing representative aggregated point features with minimal loss of information compared to the max pooling function that completely neglects non-maximum point features.

The structure of the proposed LP is shown in Figure \ref{fig:learnable_pooling}.
The module architecture of the proposed LP is inspired by the Pooling by Multihead Attention (PMA) module introduced by \citet{lee2019set_settransformer}, but designed for a more compact form.
That is, we only utilize linear transforms to match the channel size of the input point features to the desired output channel size and the multi-head attention mechanism.
Note that, in this setting, the LP is a symmetric function so that the function is permutation invariant to the points of the point cloud.

\textbf{Property 1.} \textit{The proposed learnable pooling is a symmetric function that is invariant to the permutation of the points of the point cloud}.
The proof can be found in Appendix A.1.

The key to the permutation invariant property of the LP is the use of the point-wise shared-MLP and the fact that both \textit{keys} and \textit{values} originate from the same row-permuted feature matrix.
Since both \textit{keys} and \textit{values} are row-permuted by the same permutation matrix, and the permutation matrices are orthogonal, the scaled dot-product attention mechanism becomes permutation invariant.
In addition to the theoretical proof in Appendix A.1, we also show the empirical results in Section 4 to demonstrate the similarity between the standard deviations of the PointStack with LP and PointStack with max pooling outputs for various permutations of the input points.

\section{Experiment and Discussion}
In this section, we describe the dataset, network details, and training setup used for the experiments.
Then we show and discuss the experimental results.

\subsection{Implementation Details}
\textbf{Dataset} We evaluate the proposed PointStack on two tasks, 3D shape classification and point-wise part segmentation, with three different datasets: ModelNet40 \citep{wu20153d_modelnet40}, ScanObjectNN \citep{uy2019revisiting_scanobjectnn}, and ShapeNetPart \citep{yi2016scalable_shapenetpart}.
We choose the two tasks, because they represent the two extremes of the downstream tasks widely studied for point cloud data.
That is, classification requires learning the global context of the overall point cloud, while segmentation additionally requires learning the local context of each point.
In the following experiments, the number of input points is set to 1024 for the classification task and 2048 for the segmentation task.
Note that the hardest variant of ScanObjectNN (PB\_T50\_RS) is used in the experiments, where objects are perturbed with translation, rotation, and scale transformations.

\textbf{Network} 
For all experiments, we employ four residual blocks for the feature learning backbone of the PointStack, followed by an additional task-specific head.
We set the hyperparameters for the residual blocks according to \citet{ma2022rethinking_pointmlp}.
The task-specific head is made up of only MLP blocks, where each block consists of an affine transformation, batch normalization \citep{ioffe2015batch_batchnorm}, ReLU non-linearity, and dropout \citep{srivastava2014dropout} layers.
Each head has a final affine transformation layer to match the shape of the output tensors with the task-specific requirements.
For the learnable pooling, we set the size of learnable queries to $64 \times 1024$ and $1 \times 4096$ in the single-resolution pooling and multi-resolution pooling, respectively.
Since there are four residual blocks, we use four separate learnable queries for the four level of resolutions in the single-resolution pooling.

\textbf{Training Setup} We train the networks using the PyTorch library \citep{paszke2019pytorch} on RTX 3090 GPUs. 
Networks are optimized using the SGD optimizer with a cosine annealing scheduler \citep{loshchilov2016sgdr_cosine} without the warm restart.
The initial learning rate and minimum learning rate are set to be 0.01 and 0.0001, respectively, and we incorporate label smoothing \citep{szegedy2016rethinking_labelsmoothing} to the cross-entropy loss.
We perform data augmentation by applying random translation to all datasets, and random rotation to the ScanObjectNN dataset.
For the shape classification task on ModelNet40 and ScanObjectNN, we set the maximum epoch to 300 and 200, respectively, and the batch size to 48.
For the part segmentation task on ShapeNetPart, we set the batch size to 24, and the maximum epoch to 400.

\begin{table}[t]
    \centering
    \caption{Comparison of various models on ModelNet40, ScanObjectNN, and ShapeNetPart. We show the overall accuracy (OA), class mean accuracy (mAcc), and instance mIoU (Inst. mIoU). The notation $x \pm y$ represents the mean and standard deviation of the results after multiple runs of training.}
    \begin{tabular}{lccccc}
        \toprule
         & \multicolumn{2}{c}{ModelNet40} & \multicolumn{2}{c}{ScanObjectNN} & ShapeNetPart\\
        \cmidrule(lr){2-3}
        \cmidrule(lr){4-5}
        \cmidrule(lr){6-6}
        \multicolumn{1}{c}{Model} & OA & mAcc & OA & mAcc & Inst. mIoU\\
        & (\%) & (\%) & (\%) & (\%) & (\%)\\
        \cmidrule(lr){1-6}
       
        PointCNN \citep{li2018pointcnn_pointcnn}& 92.5 & 88.1 & 78.5 & 75.1 & 86.1\\
        SpiderCNN \citep{xu2018spidercnn} & 92.4 & - & - & - & 85.3\\
        DGCNN \citep{wang2019dynamic_dgcnn} & 92.9 & 90.2 & 78.1 & 73.6 & 85.2\\
        KPConv \citep{thomas2019kpconv_kpconv} & 92.9 & - & - & - & 86.4 \\
        DRNet \citep{qiu2021dense_drnet} & 93.1 & - & 80.3 & 78.0 & 86.4\\
        GBNet \citep{qiu2021geometric_gbnet} & 93.8 & 91.0 & 80.5 & 77.8 & 85.9\\
        Simpleview \citep{goyal2021revisiting_simpleview} & 93.9 & 91.8 & 80.5 & - & -\\
        MVTN \citep{hamdi2021mvtn_mvtn} & 93.8 & \textbf{92.2} & 82.8 & - & -\\
        CurveNet \citep{xiang2021walk} & 93.8 & - & - & - & 86.8\\
        PointBERT \citep{yu2021point_pointBERT} & 93.8 & - & 83.1 & - & 85.6\\
        PRA-Net \citep{cheng2021net_pranet} & 93.7 & 91.2 & 82.1 & 79.1 & 86.3\\
        Point-MAE \citep{pang2022masked_pointMAE} & 94.0 & - & 85.2 & - & 86.1\\
        Point-TnT \citep{berg2022points_pointtnt} & 92.6 & - & 83.5 & 81.0 & -\\
        \cmidrule(lr){1-6}
        PointNet \citep{qi2017pointnet_pointnet} & 89.2 & 86.0 & 68.2 & 63.4 & 83.7\\
        PointNet++ \citep{qi2017pointnet++_pointnet++} & 90.7 & - & 77.9 & 75.4 & 85.1\\
        PointMLP \citep{ma2022rethinking_pointmlp} & \textbf{94.1} & 91.5 & 85.4 $\pm$ 0.3 & 83.9 $\pm$ 0.5 & 86.1\\
        \cmidrule(lr){1-6}
        PointStack & 93.3 & 89.6 & \textbf{86.9} $\pm$ \textbf{0.3} & \textbf{85.8} $\pm$ \textbf{0.3} & \textbf{87.2}\\
        & & &best = 87.2 & best = 86.2\\
        \bottomrule \\
    \end{tabular}
    \label{tab:main_result}
\end{table}

\subsection{Shape Classification}
We evaluate the proposed PointStack on the shape classification task with ModelNet40 and ScanObjectNN datasets.
ModelNet40 is a synthetic dataset of 40 different shape categories in the 12,311 point clouds sampled from computer-aided design (CAD) meshes.
ScanObjectNN, on the other hand, acquires point clouds from real-world object scans, thus, the samples contain background points and occlusions.
There are about 15,000 point clouds of 15 different shape categories.

Experimental results in Table \ref{tab:main_result} show that PointStack well outperforms the previous MLP-based network, PointMLP \citep{ma2022rethinking_pointmlp}, on the real-world dataset (i.e., ScanObjectNN dataset) by 1.5\% and 1.9\% for the mean OA and mean mAcc, respectively.
PointStack also outperforms other existing works such as the multiview projection-based MVTN \citep{hamdi2021mvtn_mvtn} by 4.1\%, and the transformer-based Point-TnT \citep{berg2022points_pointtnt} by 3.4\%.
Notice that, PointStack reduces the gap between the OA and mAcc performance, proving that PointStack is less biased towards certain classes compared to existing works.
The shape classification results prove that minimizing the loss of information concerning granularity and non-maximum point features through the multi-resolution feature learning and LP is beneficial for tasks that rely on the global context of the point cloud.

We notice that the overall accuracy performance of the PointStack on the synthetic dataset (i.e., ModelNet40) stands competitively at 93.3\%, which is not superior to existing works.
We speculate that the underlying cause of this issue is the significantly smaller number of training samples available in ModelNet40.
To support this speculation, we train PointStack and PointMLP on a small subset of ScanObjectNN dataset, which we discuss in more detail in Subsection 4.6.

\subsection{Part Segmentation}


We evaluate the proposed PointStack on the part segmentation task with the ShapeNetPart dataset, a synthetic dataset derived from the ShapeNet dataset.
It contains 16,881 pre-aligned point cloud shapes that can be categorized into 16 shape classes and a total of 50 segmentation classes.

\begin{figure}[ht]
    \centering
    \includegraphics[width=0.95\textwidth]{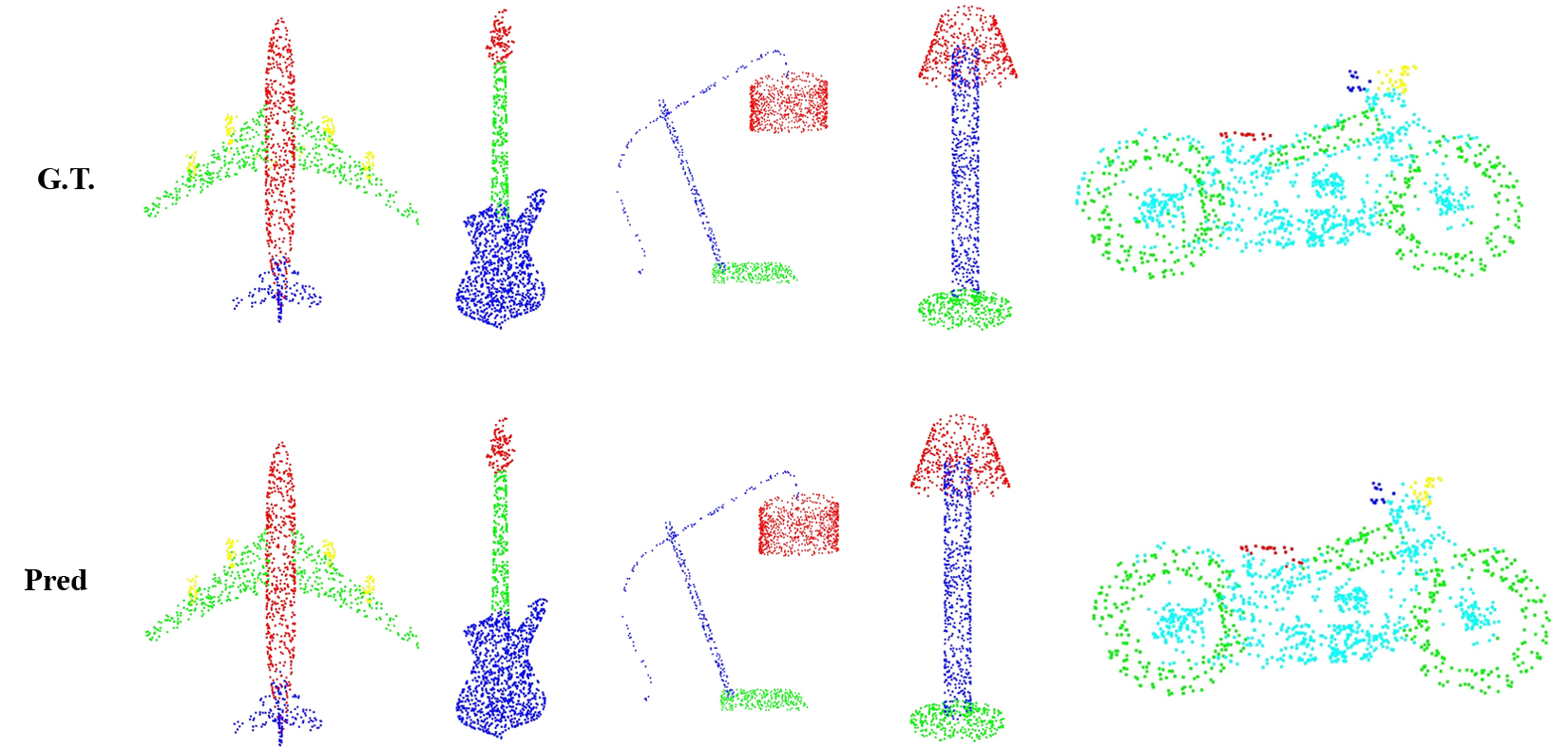}
    \caption{Visualization of the PointStack part segmentation ground truths (G.T.) and predictions (Pred). Qualitatively, the predictions are nearly identical to the ground truths.}
    \label{fig:partseg_vis}
\end{figure}

From experimental results shown in Table \ref{tab:main_result}, we observe that PointStack outperforms existing feature learning networks by at least 0.4\%.
Note that PointStack achieves such a high performance without using the voting strategy used by \citet{xiang2021walk}, where each input point cloud is randomly scaled multiple times, and the predicted logits are averaged to produce the final class prediction.
It is worth to notice that PointStack achieves such performance with a simple MLP-based network, and the best performance of existing MLP-based network \citep{ma2022rethinking_pointmlp} is 1.1\% lower than the PointStack.
The part segmentation result, especially the significant improvement from the existing MLP-based network, testifies that minimizing the loss of information concerning granularity and non-maximum point features is crucial for tasks that require both global and local contexts.
We visualize the part segmentation results in Figure \ref{fig:partseg_vis} to demonstrate the high performance of the PointStack.

\subsection{Ablation Study}
We conduct an ablation study with the ScanObjectNN dataset to investigate the effect of three major components of PointStack on the classification performance.
The three major components are multi-resolution features, LP-based single-resolution pooling, and LP-based multi-resolution pooling.

First, we investigate the effect of multi-resolution features.
We apply the max pooling function to the point features of four different resolution levels, which results in four single-resolution global feature vectors.
As in PointStack, we concatenate the four single-resolution global feature vectors and then apply another max pooling operation, which produces the multi-resolution global feature vector.
From Table \ref{tab:ablation}, we observe that incorporating multi-resolution features improve the OA and mAcc performances by 0.4\% and 0.5\%, respectively.
This result proves that preserving granularity through multi-resolution feature learning is beneficial for the classification performance of the network.

Second, we examine the effect of the Learnable Pooling (LP).
We replace the max pooling function in the single-resolution pooling process with the LP.
When the LP is used for pooling feature vectors from each level of the four resolutions, the OA and mAcc scores are further improved by 0.7\% and 0.8\%, respectively.
Subsequently, when the LP is used for the multi-resolution pooling, PointStack gains additional 0.4\% and 0.6\% performance improvements for OA and mAcc, respectively. 
This result demonstrates that appropriately utilizing information from all point features in both single-resolution and multi-resolution poolings are crucial for producing relevant representations that benefit the classification performance of the network.

Additionally, we emphasize the importance of the single-resolution LP process.
As mentioned in Section 3, the single-resolution LP enables PointStack to pool an equal number of feature vectors from each level of the \textit{m} resolutions.
In Table \ref{tab:ablation}, the performance of PointStack without single-resolution LP becomes 0.9\% lower for both OA and mAcc, in addition to the higher variance.
This result indicates that standardizing the number of feature vectors from each level of the \textit{m} resolutions is indeed crucial for the multi-resolution LP to achieve high performance.

\begin{table}[t]
    \centering
    \caption{Ablation study on the PointStack major components on the ScanObjectNN dataset. The notation $x \pm y$ represents the mean and standard deviation of the results after several runs of training.}
    \begin{tabular}{cccccc}
        \toprule
        Multi-resolution & Single-resolution & Multi-resolution & OA & mAcc \\
        Features & LPs & LP & (\%) & (\%) \\
        \cmidrule(lr){1-5}
        - & - & - & 85.4 $\pm$ 0.3 & 83.9 $\pm$ 0.5 \\
        \checkmark & - & - & 85.8 $\pm$ 0.1 & 84.4 $\pm$ 0.1  \\
        \checkmark & \checkmark & - & 86.5 $\pm$ 0.4 & 85.2 $\pm$ 0.2 \\
        \checkmark & \checkmark & \checkmark & \textbf{86.9 $\pm$ 0.3} & \textbf{85.8 $\pm$ 0.3}\\           
        \checkmark & - & \checkmark & 86.0 $\pm$ 0.7 & 84.9 $\pm$ 0.4\\
        \bottomrule \\
    \end{tabular}
    \label{tab:ablation}
\end{table}

\subsection{Permutation Invariant Property of the Learnable Pooling}
As mentioned in Section 3, the pooling function for any point cloud feature learning network should be permutation invariant.
That is, the pooling function should be capable of producing the same output even if the order of the input points is changed.

\begin{table}[t]
    \parbox{.45\linewidth}{
        \centering
        \caption{Comparison of the standard deviation of the OA ($\sigma$ OA) for shape classification on the ScanObjectNN dataset. We use ten random permutations to calculate the $\sigma$ OA values. \label{tab:pooling_variance}}
        \begin{tabular}{lc}
            \toprule
            \multicolumn{1}{c}{\textbf{Pooling Function}} & \multicolumn{1}{c}{$\sigma$ OA (\%)} \\
            \cmidrule(lr){1-2}
            Max Pooling & 0.22\\ 
            Learnable Pooling & 0.26\\
            \bottomrule \\
        \end{tabular}
    }
    \hfill
    \parbox{.45\linewidth}{
        \centering
        \caption{Performance comparison on the ScanObjectNN dataset. OA$_{F}$ and OA$_{S}$ are the classification performances when trained on the full and subset of ScanObjectNN dataset, respectively. \label{tab:scanobjectnn_limited}}
        \begin{tabular}{lc}
            \toprule
            \multicolumn{1}{c}{\textbf{Model}} & OA$_F$ (\%) $\rightarrow$ OA$_S$ (\%)\\
            \cmidrule(lr){1-2}
            PointMLP & 85.4 $\pm$ 0.3 $\rightarrow$ 73.9 $\pm$ 0.3 \\
            PointStack & 86.9 $\pm$ 0.3 $\rightarrow$ 71.7 $\pm$ 0.3 \\
            \bottomrule \\
        \end{tabular}
    }
\end{table}

To evaluate the permutation invariant property of the learnable pooling, we compare two variants of the PointStack: one with max pooling and another with the proposed learnable pooling.
Specifically, we train the two variants and evaluate the standard deviation of the OA for ten random permutations of the input points.

From Table \ref{tab:pooling_variance} we see that the network with learnable pooling has a similar standard deviation to the network with max pooling, where the standard deviation difference is only 0.04\%.
As the standard deviations are both small and similar, we confirm that the learnable pooling has the permutation invariant property similar to the max pooling.

\subsection{Limitations on the Number of Training Samples}

We observe that although PointStack achieves state-of-the-art performance on the ScanObjectNN dataset, it does not outperform existing works on the ModelNet40 dataset.
From this observation, we speculate that a potential cause of lower performance of the PointStack can be the insufficient training samples available in the ModelNet40 dataset.
The ModelNet40 dataset has 9,843 point clouds for training 40 different classes.
In comparison, the main-PB\_T50\_RS variant of the ScanObjectNN dataset has over 11,000 point clouds for training just 15 classes.

To validate the necessity of a large number of training samples, we train PointStack and PointMLP (state-of-the-art network for ModelNet40) on a small subset of the ScanObjectNN dataset.
The subset is constructed such that the number of training samples for each class matches the average number of training samples for each class in the ModelNet40.
This translates roughly to 246 samples per class.
During training, no augmentation method is applied.

As shown in Table \ref{tab:scanobjectnn_limited}, the overall accuracy of PointStack is lower than the existing MLP-based network performance, PointMLP, when the number of training samples is insufficient.
And PointStack and PointMLP show 15.2\% and 11.5\%, respectively, lower performance than when they are trained with the full ScanObjectNN dataset.
The results show the importance of sufficient training data size for PointStack to achieve the state-of-the-art performance.
One possible explanation on such a requirement is that PointStack has a larger number of trainable parameters than the existing MLP-based networks due to the multiple learnable pooling.
However, we emphasize that PointStack still achieves a competitive performance when trained with a limited number of training samples, and that modern datasets such as ScanObjectNN have sufficiently large training samples.

\section{Conclusion}
Recent point cloud feature learning networks often use aggregated point features originating from the deepest layer when performing downstream tasks.
The aggregated point features may contain high-semantic information, but there is a cost of losing information concerning granularity and non-maximum point features due to the sampling operation and max-pooling, respectively. 
In this work, we have proposed a novel MLP-based feature learning network, PointStack, where the task-specific heads are given inputs of aggregated multi-resolution point features by a generalized pooling function, learnable pooling (LP).
As a result, the aggregated point features could effectively represent both global and local contexts, and enable the network head to well comprehend the global structure and local shape details of objects in the point cloud.
Empirically, we observe that PointStack outperforms various existing feature learning networks for the shape classification and part segmentation tasks.
In the future, it is worthwhile to investigate the effectiveness of PointStack as the feature learning backbone network for other downstream tasks such as 3D object detection and shape completion.

\section*{Acknowledgment}
This work was supported by the National Research Foundation of Korea (NRF) grant funded by the Korea government (MSIT) (No. 2021R1A2C3008370).

\medskip

{
\small
\bibliography{reference}
}
\newpage
\appendix

\section{Appendix}

\subsection{Proof for Property 1}
Let \textbf{F} be the input point features matrix to the learnable pooling function $\Psi$.
Furthermore, suppose that $\textbf{Q}$, $\textbf{K}$, $\textbf{V}$, are the \textit{query}, \textit{key}, and \textit{value} matrices respectively, for a scaled dot-product attention mechanism $\Phi$.

The learnable pooling $\Psi$ can be formally defined as

\begin{equation}
    \Psi(\textbf{Q}, \textbf{F}) = \Phi(\textbf{QW}_q, \textbf{FW}_k, \textbf{FW}_v),
    \label{eq:F_init}
\end{equation}
where $\textbf{W}_q$, $\textbf{W}_k$ and $\textbf{W}_v$ are the learnable weight matrices of linear transformations for the \textit{query}, \textit{key}, and \textit{value}, respectively. 
Following the definition of scaled dot-product attention mechanism \citep{vaswani2017attention_transformer}, equation (\ref{eq:F_init}) becomes

\begin{equation}
    \Phi(\textbf{QW}_q, \textbf{FW}_k, \textbf{FW}_v) = softmax \left(\frac{\textbf{QW}_q(\textbf{FW}_k)^T}{\sqrt{d_k}}\right)\textbf{FW}_v,
    \label{eq:LP}
\end{equation}
where $d_k$ is a scaling factor proportional to the feature dimension. Consider a case where $\textbf{F}$ is row-permuted by a permutation matrix $\textbf{P}$.
Then, the learnable pooling function becomes

\begin{equation}     
    \label{eq:P_LP}
    \begin{split}
        \Psi(\textbf{Q}, \textbf{P}\textbf{F}) & = \Phi(\textbf{QW}_q, \textbf{PFW}_k, \textbf{PFW}_v)\\
            & = softmax \left(\frac{\textbf{QW}_q(\textbf{PFW}_k)^T}{\sqrt{d_k}}\right)\textbf{PFW}_v.
    \end{split}
\end{equation}
Expanding the multiplications, and considering that permutation matrix does not scale the values such that performing the permutation before or after the softmax result in the same values, we obtain

\begin{equation}
    \label{eq:P_4}
    softmax\left( \frac{\textbf{QW}_q(\textbf{P}\textbf{FW}_k)^T}{\sqrt{d_k}} \right)\textbf{P}\textbf{FW}_v = softmax \left(\frac{\textbf{QW}_q\textbf{W}_k^T\textbf{F}^T}{\sqrt{d_k}} \right)\textbf{P}^T\textbf{P}\textbf{FW}_v.
\end{equation}

Since permutation matrices are orthogonal, i.e., $\textbf{P}\textbf{P}^T = \textbf{P}^T\textbf{P}= \textbf{I}$, where $\textbf{I}$ is an identity matrix, equation (\ref{eq:P_4}) becomes
\begin{equation}
    \label{eq:P_LP_final}
    softmax \left(\frac{\textbf{QW}_q\textbf{W}_k^T\textbf{F}^T}{\sqrt{d_k}} \right)\textbf{P}^T\textbf{P}\textbf{FW}_v = softmax \left(\frac{\textbf{QW}_q(\textbf{FW}_k)^T}{\sqrt{d_k}} \right)\textbf{FW}_v.
\end{equation}

Since the right hand side of equation (\ref{eq:P_LP_final}) is equal to the right hand side of equation (\ref{eq:LP}), we prove that $\Psi(\textbf{Q}, \textbf{F}) = \Psi(\textbf{Q}, \textbf{PF})$ and Property 1 in subsection 3.2 holds.

\end{document}